\documentclass[10pt,twocolumn,letterpaper]{article}

\usepackage[pagenumbers]{wacv} % To force page numbers, e.g. for an arXiv version

\usepackage{graphicx}
\usepackage{amsmath}
\usepackage{amssymb}
\usepackage{booktabs}
\usepackage{multirow}

\usepackage[pagebackref,breaklinks,colorlinks]{hyperref}

\usepackage[capitalize]{cleveref}
\crefname{section}{Sec.}{Secs.}
\Crefname{section}{Section}{Sections}
\Crefname{table}{Table}{Tables}
\crefname{table}{Tab.}{Tabs.}

\def\wacvPaperID{2348} % *** Enter the WACV Paper ID here
\def\confName{WACV}
\def\confYear{2024}

\usepackage{xcolor}

\usepackage{array}
\usepackage{url}
\usepackage{booktabs}

\usepackage{pifont}
\newcommand{\cmark}{\ding{51}}%
\newcommand{\xmark}{\ding{55}}%

\usepackage[accsupp]{axessibility}

\begin{document}

\title{PsyMo: A Dataset for Estimating Self-Reported Psychological Traits from Gait}
\author{Adrian Cosma\\
University Politehnica of Bucharest\\
Bucharest, Romania\\
{\tt\small cosma.i.adrian@gmail.com}
\and
Emilian Radoi\\
University Politehnica of Bucharest\\
Bucharest, Romania\\
{\tt\small emilian.radoi@upb.ro}
}

\maketitle
% \author{\IEEEauthorblockN{Adrian Cosma}
% \IEEEauthorblockA{University Politehnica of Bucharest\\
% Bucharest, Romania\\
% Email: cosma.i.adrian@gmail.com}
% \and
% \IEEEauthorblockN{Emilian Radoi}
% \IEEEauthorblockA{University Politehnica of Bucharest\\
% Bucharest, Romania\\
% Email: emilian.radoi@upb.ro}
% }

% \maketitle

\begin{abstract}
Psychological trait estimation from external factors such as movement and appearance is a challenging and long-standing problem in psychology, and is principally based on the psychological theory of embodiment. To date, attempts to tackle this problem have utilized private small-scale datasets with intrusive body-attached sensors. Potential applications of an automated system for psychological trait estimation include estimation of occupational fatigue and psychology, and marketing and advertisement. In this work, we propose PsyMo (\textbf{Psy}chological traits from \textbf{Mo}tion), a novel, multi-purpose and multi-modal dataset for exploring psychological cues manifested in walking patterns. We gathered walking sequences from 312 subjects in 7 different walking variations and 6 camera angles. In conjunction with walking sequences, participants filled in 6 psychological questionnaires, totaling 17 psychometric attributes related to personality, self-esteem, fatigue, aggressiveness and mental health. We propose two evaluation protocols for psychological trait estimation. Alongside the estimation of self-reported psychological traits from gait, the dataset can be used as a drop-in replacement to benchmark methods for gait recognition. We anonymize all cues related to the identity of the subjects and publicly release only silhouettes, 2D / 3D human skeletons and 3D SMPL human meshes.
\end{abstract}

\section{Introduction}

How we move and behave in the physical space is intrinsically tied to our psychological workings. Besides each individual's muscle developments and the influence of extrinsic factors, walking (or gait) is influenced by gender \cite{catruna2021face}, emotions \cite{bhattacharya2020take}, personality traits \cite{Satchell2017}, and mental distress \cite{fang2019depression}. Walking is an action performed ubiquitously by healthy individuals. So far, in the domain of AI-powered gait analysis, significant attention has been dedicated to gait recognition \cite{gei-original,10.1007/978-3-642-15549-9_19,chao2019gaitset,fan2020gaitpart,9324873,10.1007/978-3-319-69923-3_51,An2018ImprovingGR}, but with little focus being dedicated to exploring the manifestations of psychological traits in gait \cite{10.3389/frobt.2021.749274}. 

While there has been research in studying personality manifestations in video \cite{liu2016analyzing,celiktutan2015computational,fang2016personality,rahbar2015predicting}, walking is left largely unexplored. 
Nonetheless, some studies \cite{Satchell2017,Sun2017-aq,9399002} separately confirm significant differences in gait between individuals with different personalities, levels of aggression, depression and self-esteem. Such works are based on the assumption that psychological experience manifests itself into behaviour, in line with theories of embodiment \cite{michalak2009embodiment,zatti2015embodied}. Embodiment suggests a biofeedback influence loop between the psyche and physical gestures / actions. Actions, gestures and posture can have a clear impact on thought \cite{Goldin-Meadow2010-vk}, memory and recall \cite{Michalak2014-bd}, and mood \cite{michalak2009embodiment,posture-mood}. In this context, currently, there is no open dataset for exploring the embodied manifestation of psychological traits in gait.

We propose PsyMo (\textbf{Psy}chological traits from \textbf{Mo}tion), a multi-purpose gait database containing 312 subjects walking under multiple viewpoints and walking variations, annotated with self-assessed psychological traits from 6 psychological questionnaires: Big Five Index \cite{john1999big}, Rosenberg Self-Esteem \cite{Rosenberg+2015}, Buss-Perry Aggression \cite{buss1992aggression}, Occupational Fatigue Exhaustion/Recovery Scale \cite{winwood2005development}, Depression Anxiety Stress Scale \cite{lovibond1995}, and the General Health Questionnaire \cite{goldberg1970psychiatric}. Across all subscales and factors, PsyMo contains 17 psychological traits. Alongside self-assessed psychological traits, participants also submitted their age, gender, height and weight. PsyMo is the first public dataset of its kind and pertains to the set of "controlled" gait datasets used for benchmarking gait analysis tasks under various walking variations, similar to CASIA-B \cite{yu2006framework} and FVG\cite{fvg}.

PsyMo covers 7 walking variations: normal walking, changing clothes, slow walking, fast walking, walking with a bag, and two additional dual-tasks (i.e. walking while performing a cognitive task \cite{9184903}), largely ignored in current datasets: walking while texting and walking while talking on the phone. The walks are captured using 3 synchronized consumer video surveillance cameras (Tapo C200) to mimic adverse conditions (i.e. low-fps, fish-eye distortion, low-resolution) present in real-world surveillance scenarios, for a total of 6 viewpoints (which include the round-trips). In total, we gathered 14,976 walking sequences.

We provide 2D / 3D human poses, silhouettes, and 3D human meshes for gait processing. We extracted appearance-based silhouettes using instance segmentation with a pretrained Hybrid Task Cascade (HTC) model \cite{chen2019hybrid}, and 2D skeletons using AlphaPose \cite{li2018crowdpose}, a state-of-the-art model for pose estimation. Further, we used CLIFF \cite{li2022cliff} to estimate 3D human pose and 3D meshes in the form of parametric SMPL predictions. The dataset is fully anonymized, and subjects gave explicit and informed consent for processing and distributing it. We do not release raw videos, as they can lead to identifying the subjects in our study. We release only processed gait information in the form of silhouettes, 2D / 3D skeletons and 3D human meshes. 

PsyMo is primarily intended as a rich resource for exploring psychological manifestations into walking patterns, allowing for an interdisciplinary study into human behaviour from both the artificial intelligence and psychology research communities. We propose several evaluation procedures for the estimation of psychological traits from gait. Furthermore, due to its size and number of walking variations, PsyMo can also be used as a benchmark dataset for standard gait recognition, in addition to / as a more diverse drop-in replacement to existing gait benchmark datasets \cite{yu2006framework,fvg,segundo2023long}. For this purpose, we also proposed several evaluation protocols for gait recognition.

\section{Related Work}
\paragraph{Psychological Cues from Behaviour}
There have been several attempts and datasets \cite{7736040,junior2019first} for personality detection from external body movement and nonverbal behaviour in literature. Subramanian et al. \cite{7736040} proposed ASCERTAIN, a multimodal dataset for personality and affect recognition, in terms of Big Five personality traits. The dataset contains electroencephalogram (EEG), electrocardiogram (ECG), galvanic skin response (GSR) and facial activity readings for 58 subjects while viewing movie clips. The authors show that they can achieve above-chance recognition accuracy in emotional videos. This work provides evidence that personality affects physiological responses in the body. A large body of work is dedicated to \textit{apparent personality}, based on human faces, body postures and behaviours \cite{junior2019first}. Different from self-reported personality, apparent personality is based on external behavioural manifestations of a subject. In this area, numerous datasets are proposed, from multimodal (MHHRI \cite{celiktutan2017multimodal}, ChaLearn \cite{escalante2017design}, SEMAINE \cite{mckeown2011semaine}) to audiovisual (ELEA \cite{sanchez2011nonverbal}, Youtube Vlog \cite{biel2010voices,biel2012youtube,biel2010vlogcast}). Approaches generally make use of static face images \cite{liu2016analyzing}, videos \cite{celiktutan2015computational}, audiovisual \cite{fang2016personality} cues and multimodal \cite{rahbar2015predicting} features.  

Regarding gait analysis, Satchell et al. \cite{Satchell2017} showed in a narrow study that there are statistically significant correlations between gait features such as speed, cadence and limb range of motion and Big Five personality traits. Similarly, such correlations are found in terms of aggression (using the Buss-Perry Aggression scale). These findings motivate our work to explore automatic estimation of personality traits and signs of mental distress from walking in more realistic settings. Works in the area of depression detection and signs of mental distress from gait \cite{fang2019depression,9399002,yang2022data} generally make use of the PHQ-9 \cite{Kroenke2001-qv} for measuring depression or the GAD-7 \cite{Spitzer2006-yv} scale for measuring anxiety. However, the authors do not release the datasets to the research community. For instance, Fang et al. \cite{fang2019depression} extracted skeleton sequences of walking students using a Kinect sensor and computed several hand-crafted features for the gait data, such as walking speed, arm swing, stride length, and head movement. Using classical machine learning classifiers such as SVM and Random Forest, they obtained good results on depression detection. Sun et al. \cite{Sun2017-aq} obtained statistically significant correlations between self-esteem measured by the Rosenberg scale \cite{Rosenberg+2015} and participants' gait. The authors collected a corpus of 178 graduate students to perform their experiments and estimated the individual skeletons using a Kinect sensor. Alongside cues for personality and mental distress cues, there are some works which tackle estimation of emotions from behaviour, and in particular gait \cite{bhattacharya2020take,chiu2018emotion}.

\paragraph{Gait Analysis}
Current works in gait analysis from video focus on identity recognition, regarding gait as a unique biometric fingerprint able to identify subjects \cite{yu2006framework,zhu2021gait,chao2019gaitset,fan2020gaitpart,cosma2021wildgait}. Gait recognition is primarily performed on benchmark datasets \cite{yu2006framework,hofmann2014tum,Xu_CVA2017,fvg}, collected in controlled scenarios, most prominently CASIA-B \cite{yu2006framework}. CASIA-B is one of the first medium-scale datasets used for training and evaluating the performance of models in various walking variations (normal walking, change in clothing, carrying conditions) and different viewpoints. It contains 126 subjects and 11 viewpoints. However, one limitation of CASIA-B is its lack of size, thus researchers tend to use it as a benchmark set, and train models on larger and more diverse sets such as GREW\cite{zhu2021gait}, Gait3D \cite{zheng2022gait} or DenseGait \cite{cosma22gaitformer}. Most current datasets are not annotated beyond subject identity, with only some providing annotations for age and gender \cite{Xu_CVA2017}. FVG \cite{fvg} is a similar controlled gait dataset, containing 226 individuals walking towards the camera. In addition to the walking variations present in CASIA-B, FVG also contains walking speed and the passage of time as confounding factors. GREW, Gait3D, and DenseGait are another class of gait databases, which are collected "in the wild", with no clear delimitation between walking variations. They represent a more challenging scenario for gait recognition, with orders of magnitude more identities and with real-world walking variations.

Approaches for gait analysis typically are classified into appearance-based and model-based. Appearance-based methods more recently make use of silhouettes \cite{gei-original,10.1007/978-3-642-15549-9_19,chao2019gaitset,fan2020gaitpart} extracted either from background subtraction methods or from a pretrained instance segmentation model. Approaches typically use either a compressed form of the silhouette sequence through a Gait Energy Image (GEI) \cite{gei-original}, or process the silhouettes as they are \cite{chao2019gaitset,fan2020gaitpart}. Model-based approaches, on the other hand, make use of a model of the human body. More recently, such approaches entail the use of skeletons automatically extracted from pretrained pose estimation models \cite{cosma2021wildgait}, and processing them with architectures borrowed from skeleton action recognition, such as a CNN \cite{9324873}, an LSTM \cite{10.1007/978-3-319-69923-3_51,An2018ImprovingGR} or graph methods such as ST-GCN \cite{li2020jointsgait,9721551}. Furthermore, many recent works are starting to take advantage of the increasingly accurate predictions of the 3D human body model (e.g., SMPL\cite{loper2015smpl}) and incorporate 3D information in their pipelines \cite{zheng2022gait,segundo2023long,zhu2023gait}.

\section{The PsyMo Dataset}
\label{sec:dataset}
%The questionnaires were delivered using Google Forms\footnote{\url{https://forms.google.com/}} and were filled in remotely. 

To gather our dataset, we asked volunteers to fill in 7 psychological questionnaires and personal information about their body composition: age, gender, height and weight. After submitting the questionnaires, participants were required to have their walk captured in various conditions. All participants are physically healthy and do not report any injury or disability affecting their walking. Participants were dressed in everyday clothes. In total, 312 subjects participated in our study, mostly university students. All participants were explicitly informed about how their data is being processed and that the anonymized dataset is made available to the research community at large. Subsequently, all participants gave explicit consent to be part of this study before filling in the questionnaires. Moreover, this study was approved by the Ethics Review Board at the University Politehnica of Bucharest (AC01/01.10.2021). The dataset is made publicly available \footnote{PsyMo Dataset Link: \url{https://bit.ly/3Q91ypD}}. The dataset is protected by the CC-BY-NC-ND\footnote{\url{https://creativecommons.org/licenses/by-nc-nd/4.0/legalcode}} License.

\subsection{Collecting Psychometric Data}
As opposed to other similar works \cite{fang2019depression,yang2022data}, we have a more diverse set of questionnaires, which correspond to indicators of personality (BigFive, Rosenberg Self-Esteem), fatigue (OFER), aggressiveness (Buss-Perry Aggression Questionnaire) and mental health (General Health Questionnaire, Depression, Anxiety and Stress Scale). For the participants' convenience, we chose the short version of the questionnaires when possible, resulting in a total completion time of approximately 30 minutes for all questionnaires. Since all our subjects are native Romanians, we used translations for each questionnaire when available. In case a translation is not available, we performed the translations with the help of a psychologist. Before capturing the walking videos, each participant was required to fill in the questionnaires. Of the 312 participants, 113 were female and 199 were male, with an average age of 21.9 years (SD = 2.18). The average weight for the participants is 70.5kg (SD = 15.7), with an average height of 174.8cm (SD = 8.9), corresponding to an average BMI of 22.87 (SD = 3.9). 
% Figure \ref{fig:gender-distribution} showcases the distribution for each attribute. 
We briefly describe the motivation and technical details of each psychological questionnaire.

% \begin{figure*}[hbt!]
%     \centering
%     \includegraphics[width=0.80\textwidth]{images/gender.png}
%     \caption{Distributions of personal attributes (weight, height, BMI) for all participants, split by gender.}
%     \label{fig:gender-distribution}
% \end{figure*}

\noindent\textbf{Big Five Inventory (BFI)}
The Big Five Inventory \cite{john1999big} is a widely used self-report scale for grouping personality traits. It is comprised of 44 items rated on a five-point Likert scale \cite{vogt2011dictionary} from 1 ("disagree a lot") to 5 ("agree a lot"). Each item is a statement that is a verbal descriptor for a person's behaviour:  "I see myself as someone who ..." \textit{is talkative / gets nervous easily / is reserved / tends to be lazy etc}. Through factor analysis, the OCEAN (openness to experience, conscientiousness, extraversion, agreeableness, neuroticism) five factor model is revealed, describing the subject's personality. We chose the BFI as one of the questionnaires to explore the ways in which personality cues are present in human movement in the form of walking \cite{Robinson2021}. Previous works \cite{Satchell2017} have found strong correlations between big five personality and various aspects of gait (e.g. limb ranges of motion, gait speed). There are numerous applications of personality estimation from movement in real-life scenarios: marketing and advertisement, surveillance, pedestrian analysis and retail video analytics. BFI has been validated on the Romanian population \cite{david2015national}, and we used one of the available Romanian translations. After completion, we obtain the following scores for the personality traits: Openness - M = 36.68, SD = 5.37, Conscientiousness - M = 31.02, SD = 5.81, Extraversion - M = 25.29, SD = 5.48, Agreeableness - M = 33.45, SD = 4.71, Neuroticism - M = 22.40, SD = 5.59. BFI has no cutoff points for the final scores of each personality traits, but rather must be compared to a standard population. As such, we compute the score quantiles and for each personality trait (i.e. OCEAN) we obtain 4 ordinal classes: Low, Normal/Low, Normal/High and High.

\noindent\textbf{Rosenberg Self-Esteem (RSE)}
Similar to the Big-Five Inventory, self-esteem is an important psychological trait, referring to the subjects' subjective evaluation of self-worth. Self-esteem is correlated with other psychological attributes: lower levels of self-esteem are associated with increased levels of depression and anxiety \cite{van2014implicit}, and is strongly inversely correlated with neuroticism \cite{amirazodi2011personality}. Nevertheless, we chose to collect a separate and explicit self-esteem score for more reliable estimation from gait and posture. Other works \cite{Sun2017-aq} report a significant correlation between some gait features and Rosenberg self-esteem scores.

The Rosenberg Self-Esteem Scale \cite{Rosenberg+2015} contains 10 items that measure global self-worth, with both positive and negative attitudes towards the self. Each item is rated using a 4-point Likert scale from 0 ("strongly disagree") to 3 ("strongly agree"), with scores ranging from 0 to 30. The scale is validated on the Romanian population \cite{robu2015scala}, and we use one of the available translations. 

The Rosenberg Self-Esteem Scale is uni-dimensional, and after completion, we obtain an average score of 30.83 (SD = 5.50). Scores are recommended to be kept in a continuous scale \cite{Rosenberg+2015}, but for uniformity with the other questionnaires, we compute 3 ordinal classes (Low, Normal, High) using known thresholds.

\noindent\textbf{Buss-Perry Aggression Questionnaire (BPAQ)}
The Buss-Perry Aggression Questionnaire \cite{buss1992aggression} contains 29 items that measure physical aggression (9 items), verbal aggression (5 items), anger (8 items) and hostility (8 items). Items are scored using a 5 point Likert scale, ranging from 1 ("extremely uncharacteristic of me") to 5 ("extremely characteristic of me"). Works such as the one from Satchell et al. \cite{Satchell2017} found correlations between gait characteristics and factors of the BPAQ, indicating that automatic estimation is feasible. Estimating aggression from movement is crucial in safety situations by monitoring abnormal behaviour, i.e. in airports and train stations. After completion, we obtained the following scores for each factor: physical aggression - M = 24.65, SD = 3.71, verbal aggression - M = 11.55, SD = 3.51, anger M = 15.77, SD = 4.60, hostility - M = 17.83, SD = 4.52. Ordinal classes are obtained by quantization through quantiles. We obtain 4 classes: Low, Normal/Low, Normal/High and High.

\noindent\textbf{Occupational Fatigue Exhaustion/Recovery Scale (OFER)}
Walking is a fundamental part of industrial workers in areas such as mining, construction and distribution centres. Manifestations of fatigue in walking are correlated with fatigue in other physical tasks and unintrusive estimation of workplace fatigue offers clear advantages for mitigation of workplace risks and monitoring of worker performance. To measure occupational fatigue, we used the Occupational Fatigue Exhaustion/Recovery Scale \cite{winwood2005development}, which contains 15 items, split into 3 factors: chronic fatigue (5 items), acute fatigue (5 items) and recovery (5 items). It is mainly used as a psychometric indicator of inter-shift recovery rate and persistent occupational fatigue in the workplace. 
Each item in the OFER scale is scored using a 7-point Likert scale, ranging from 0 ("strongly disagree") to 6 ("strongly agree"). We translated each item with the help of a psychologist. Scores for each factor are calculated separately (item scores / 30 x 100), and the authors of the OFER questionnaire provide score thresholds for each factor. After completion, we obtain the following average scores: chronic fatigue - M = 38.08, SD = 21.30, acute fatigue - M = 40.20, SD = 19.42, recovery - M = 49.55, SD = 17.7. As such, the final ordinal classes are divided into Low, Moderate/Low, Moderate/High and High. 

\noindent\textbf{Depression, Anxiety and Stress Scale (DASS-21)}
The Depression, Anxiety and Stress Scale \cite{lovibond1995} is a set of three self-report scales measuring depression, anxiety and stress, for a total of 42 items. DASS-21 is a widely accepted questionnaire and is suitable for clinical settings to assist in diagnosis, as well as non-clinical settings for screening mental health problems. The DASS scale measures signs of mental disorders, which could prove valuable if estimated from movement and gait in remotely monitoring the mental state of individuals in scenarios of assisted living, worker performance monitoring and healthcare institutions. Detection of mental health manifestations from movement is in line with the theory of embodiment \cite{michalak2009embodiment}, which suggests a reciprocal relationship between the psyche and movement. Similar to us, Fang et al. \cite{fang2019depression} constructed a system for depression detection from gait in Chinese university students, making use of PHQ-9 depression questionnaire \cite{Kroenke2001-qv}, and obtained significant detection accuracy. However, we chose DASS-21 since it contains three subscales, including depression. Each item is rated on a 4-point Likert scale, ranging from 0 ("did not apply to me") to 3 ("applied to me very much, or most of the time"). It contains statements regarding negative emotional states that apply to an individual thinking about last week, before the moment of taking the inventory. In our work, we chose the short version with 21 items (DASS-21) for a faster completion time, without affecting the final scores. For each subscale, scores are in the range of 0 and 42. The authors provide translations in many languages, and we used the Romanian translation in our study. After completion, we obtained the following scores for each subscale: depression - M = 24.86, SD = 9.63, anxiety - M = 26.10, SD = 8.67, stress - M = 26.68, SD = 8.26. Ordinal classes for each subscale are computed based on the threshold values provided by the authors: Normal, Mild, Moderate, Severe and Extremely Severe.

\noindent\textbf{General Health Questionnaire (GHQ)}

%%%%%%%%%%%%%%
\begin{figure*}[hbt!]
    \centering
    \includegraphics[width=0.85\textwidth]{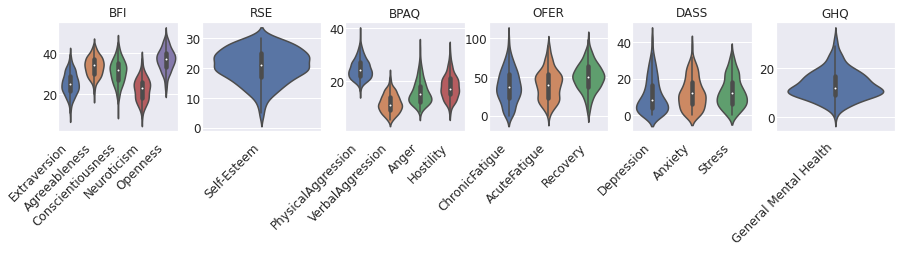}
    \caption{Violin plots for the distributions of questionnaire responses for all questionnaires and their corresponding factors / subscales. }
    \label{fig:explore-qs}
\end{figure*}
%%%%%%%%%%%%%%

The General Health Questionnaire \cite{goldberg1970psychiatric} is a widely used scale for screening signs of mental distress and minor psychiatric disorders in the general population. The questionnaire is suitable for short term psychiatric disorders, but not long-standing mental health issues, and is usually regarded as a first step to inform further intervention. In our study, we used the short form (GHQ-12) for a faster completion time, with reliable results. The questionnaire is not scaled, and offers a single score to the individuals, assessing whether the subject's current mental state differs from "normal". Each of the 12 items is rated on a 4-point Likert scale, ranging from 0 ("not at all") to 3 ("more than usual"), with total scores ranging from 0 to 36. The scale is validated on the Romanian population \cite{ghq12}, and we used an available translation for the Romanian language \cite{ghq12}. Similar to DASS-21, GHQ-12 measures signs of mental health issues, with the same potential applications in real-world unintrusive estimation from movement. After completion, we obtained an average score of 13.27 (SD = 5.76). We used the available thresholds to obtain 3 ordinal classes: Typical, Minor Distress and Major Distress.

\begin{figure}[hbt!]
    \centering
    \includegraphics[width=0.85\linewidth]{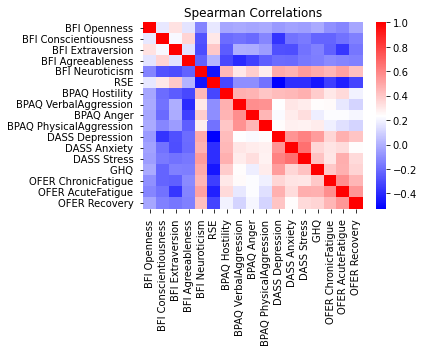}
    \caption{Spearman correlations between questionnaire scores.}
    \label{fig:correlations-all}
\end{figure}

\noindent\textbf{\textbf{Discussion}}
Figure \ref{fig:explore-qs} showcases the distribution of responses for each questionnaire and factors / subscales. While our population is mostly comprised of university students, there are no severe imbalances in the responses. Moreover, our questionnaire results are reliable and are validated by correlations found in psychology research. For instance, in our results, the Neuroticism factor in the Big Five Index is strongly correlated with other traits that we studied (Spearman's correlation p $<$ 0.0001), such as RSE, OFER and DASS scales. Neuroticism refers to a tendency to respond with negative emotion to unfavourable events such as frustration, loss and threat. Neurotic individuals are more likely to have lower self-esteem, are more tired and are more likely to experience signs of mental distress (depression, anxiety) \cite{Lahey2009}. Moreover, there are strong correlations between fatigue and anxiety / depression \cite{serrano2021work}, also found in our dataset (Spearman's correlation p $<$ 0.0001). Self-esteem is inversely correlated with depression and anxiety \cite{van2014implicit}, another property also found in our results (Spearman's correlation p $<$ 0.0001). Figure \ref{fig:correlations-all} showcases a heatmap of the correlation coefficients between each of the 17 traits in our dataset.

\subsection{Capturing Walks}
\label{sec:walks}

\begin{figure*}[hbt!]
    \centering
    \includegraphics[width=0.85\textwidth]{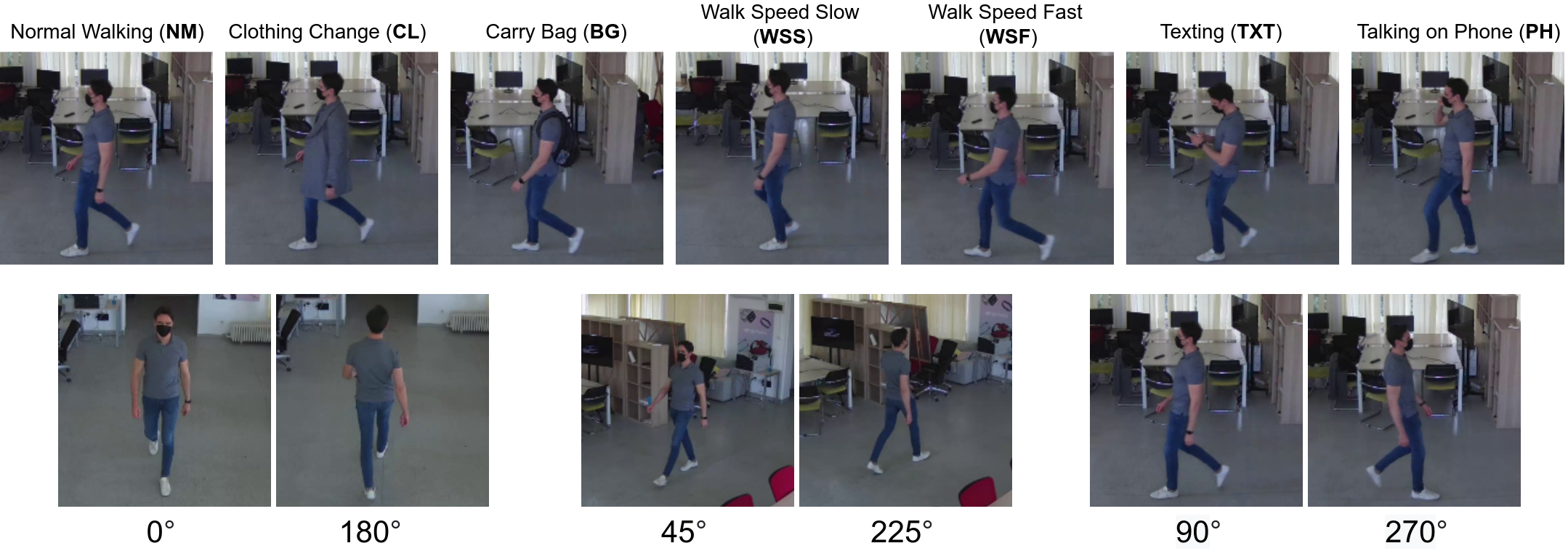}
    \caption{Walking variations and viewpoints for the gait sequences in our dataset. Participants walked in 7 variations (NM, CL, BG, WSS, WSF, TXT and PH) and their walks were captured from 6 viewpoints.}
    \label{fig:psymo-variations}
\end{figure*}

\begin{figure}[hbt!]
    \centering
    \includegraphics[width=0.90\linewidth]{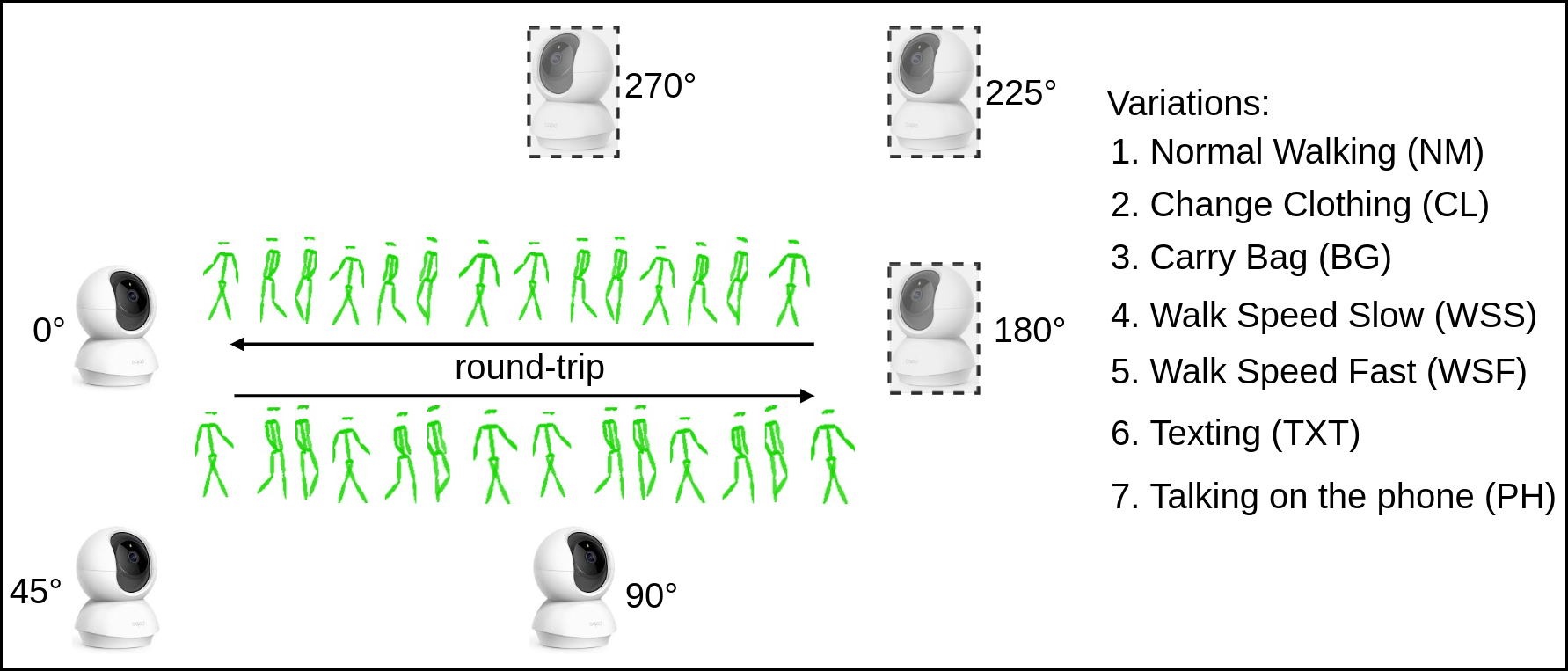}
    \caption{Setup for capturing walking sequences. We employed three synchronized Tapo C200 consumer cameras, at three angles: 0$^{\circ}$, 45$^{\circ}$ and 90$^{\circ}$. Since subjects perform a round-trip, we have an additional 3 virtual angles of 180$^{\circ}$, 225$^{\circ}$ and 270$^{\circ}$. Subjects walk a total of 16 times across variations, corresponding to 48 walks from all viewpoints and variations.}
    \label{fig:capturing-walks}
\end{figure}

\noindent\textbf{Experimental Setup}
Figure \ref{fig:capturing-walks} showcases our gait capturing setup. Instead of capturing walks with a high-definition, high framerate camera, we opted for more realistic hardware to simulate real-world surveillance conditions. We chose three Tapo C200 consumer surveillance cameras, for their widespread availability, ease of use and ease of interaction via a web API. The cameras record video at 1920x1080 resolution, with a framerate of 16 frames per second and were positioned at 0$^{\circ}$, 45$^{\circ}$ and 90$^{\circ}$ angles from the walking area. The walking area is 8 meters long, which allows for 5--6 full gait cycles \cite{00004623-196446020-00009}, depending on the height of the person. In the laboratory there was only one researcher present alongside the subject at any given time, to avoid influencing the subject's walking. Friesen et al. \cite{Friesen2020} showed a positive correlation between the number of researchers present in the laboratory and changes in gait speed and cadence, while not being affected by the participants' personalities.

In Figure \ref{fig:psymo-variations} we show all the walking variations and viewpoints. Alongside the standard array of covariates \cite{yu2006framework,ouisir,fvg} which includes normal walking (\textbf{NM}), changing clothing (\textbf{CL}), carry bag (\textbf{CB}) and walking speed (in our case walking speed slow - \textbf{WSS} and walking speed fast - \textbf{WSF}), we also introduce two covariates that are often ignored in the literature, but nevertheless appear frequently in the wild, dual-tasks, which have been shown to produce walking variability in adults \cite{9184903}. We introduce walking while texting (\textbf{TXT}) and walking while talking on the phone (\textbf{PH}). In total, our dataset contains 7 walking variations. For each variation, participants were asked to walk a round-trip, in a comfortable manner. For the normal walking variation, participants walked two round-trips. As such, there are a total of 48 walking sequences per participant, from 7 variations and 6 viewpoints ($3\times4$ NM $+$ $3\times2\times6$ other variations).

\begin{figure}[hbt!]
    \centering
    \includegraphics[width=0.85\linewidth]{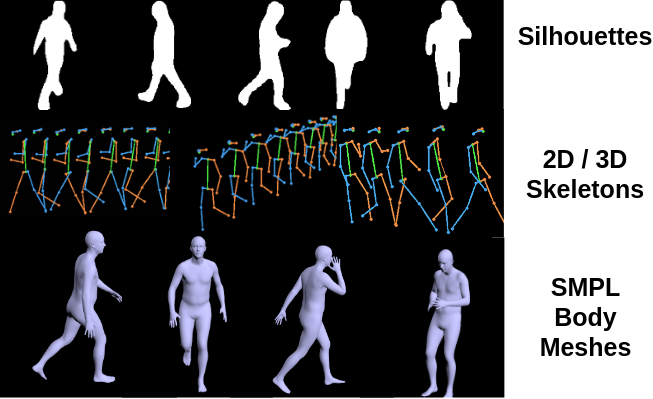}
    \caption{We publicly release only processed silhouettes, 2D / 3D skeleton sequences and 3D human meshes / SMPL. Raw RGB videos are not made available to protect the privacy of the subjects in our study.}
    \label{fig:data-format}
\end{figure}

\noindent\textbf{Data Processing} 
In order to preserve the anonymity of the participants, we do not release the full RGB videos, and instead choose to release only high-level features related to gait: silhouettes, 2D / 3D human poses and 3D human meshes (Figure \ref{fig:data-format}). We chose state-of-the-art pretrained models that have especially good performance in "in the wild" scenarios, which are arguably more complex than our laboratory conditions: more people, filmed at a distance, with detection overlap. As such, we make a reasonable assumption that the resulting semantic data (skeletons / silhouettes) is of high quality. However, we are committed to maintaining the dataset's quality, and we intend to update the semantic features with improved models over time, if necessary. We extracted silhouettes using a pretrained Hybrid Task Cascade (HTC) \cite{chen2019hybrid} model, with a ResNet50 \cite{he2016deep} backbone, trained on MSCOCO \cite{lin2014microsoft}. Silhouettes are widely used as an input representation for appearance-based gait analysis \cite{gei-original,10.1007/978-3-642-15549-9_19,5522296,chao2019gaitset,fan2020gaitpart,lin2021gait}. After processing the videos, silhouettes are centred in the image and downsized to 128x128 pixels. Furthermore, for convenience, we computed a Gait Energy Image (GEI) \cite{gei-original} for each walk. 

Skeleton sequences are also used in modern, model-based approaches for gait analysis \cite{9324873,10.1007/978-3-319-69923-3_51,An2018ImprovingGR,li2020jointsgait,9721551}. Currently, methods using skeletons sequences are outperformed by silhouette-based methods, since the extraction quality severely affects downstream computation. It is not clear whether silhouette-based methods are processing gait strictly in the form of movement, or also rely on the limited appearance information contained by the silhouette. However, skeletons allow for a more low-level analysis of walking patterns by analyzing the movement and correlation between the joints and individual body parts. We extracted 2D skeletons using AlphaPose \cite{li2018crowdpose}, a state of the art, multi-person 2D pose estimation model. AlphaPose extracts skeletons in the COCO pose format, consisting of 18 joints, with (x, y) coordinates in the image space and a confidence score for each joint, which roughly measures the detection quality. We do not do any further processing on the poses and release them as is, in JSON format.

% Further, many gait analysis works are increasingly using 3D human meshes in their processing pipeline to aid recognition and disambiguate 2D signals \cite{zheng2022gait,segundo2023long,zhu2023gait}. We extracted 3D human poses and triangle meshes (in the form of SMPL \cite{loper2015smpl} models) using CLIFF \cite{li2022cliff}, a global aware model, that has shown good results in 3D mesh extraction in the wild. For each walking video, we release the 3D pose (24 joints with x, y, z coordinates, centred at the pelvis) and SMPL parameters (10-dimensional vector) without any further modification.

Further, many gait analysis works are increasingly using 3D human meshes in their processing pipeline to aid recognition and disambiguate 2D signals \cite{zheng2022gait,segundo2023long,zhu2023gait}. We extracted 3D human poses and triangle meshes (in the form of SMPL \cite{loper2015smpl} models) using CLIFF \cite{li2022cliff}, a global context-aware model, that has shown good results in 3D mesh extraction in the wild. For each walking video, we release the 3D pose (24 joints with x, y, z coordinates, centred at the pelvis) and SMPL parameters (10-dimensional vector) without any further modification.

\subsection{Evaluation Protocols}
In all protocols, subjects with IDs from 0-250 should be used for training, and evaluation ought to be performed on subjects with IDs from 251-312, corresponding to an 80:20 training-evaluation split. 

\noindent\textbf{Psychological Traits Estimation} For psychological traits estimation on PsyMo, we propose two evaluation methodologies: (i) \textit{run-level} and (ii) \textit{subject-level}. For \textit{run-level} evaluation, the model performance is evaluated for each walking sequence, irrespective of the subject. Performance in terms of precision, recall, weighted F$_1$ score should be reported for all combinations of walking variations and viewpoints. This protocol is similar to a typical gait classification task (i.e. input walking sequence, output classes). For \textit{subject-level} evaluation, the goal is to correctly identify the psychometric attributes for each subject, considering all available variations or runs. For instance, a naive baseline for subject-level evaluation is to use the run-level model and report the majority predicted classes for a questionnaire for all the runs of a subject. Methods for subject-level evaluation may consider the identity of the subject to be known at test time - a scenario possible in the real-world, as a part of a larger pipeline for gait recognition and classification. The same metrics as in run-level evaluation should be used.

% \noindent\textbf{Gait Recognition}
% Larger "in the wild" datasets such as GREW and Gait3D are typically not annotated with information of walking covariates, and instead focus on benchmarking models' ability to scale to many identities. While these datasets are more realistic from both hardware and environment perspective, it is hard to accurately quantify failure cases (e.g. model fails on a particular walking variation). Smaller, more controlled datasets such as CASIA-B and FVG have a smaller number of identities, but with fixed number of covariates. PsyMo brings together the best of both worlds by replicating real-world video hardware for capturing walks and annotating each run with more variations and with metadata information about each subject (gender, age, weight, height). In Table \ref{tab:dataset-comparison} we showcase a comparison of PsyMo to other popular datasets for the gait recognition task. As such, PsyMo can be used as an additional benchmark and drop-in replacement of older datasets (e.g., CASIA-B \cite{yu2006framework}, de facto standard) for evaluating gait recognition models. 

\noindent\textbf{Gait Recognition}
Larger "in the wild" datasets such as GREW and Gait3D are typically not annotated with information on walking covariates, and instead focus on benchmarking models' ability to scale to many identities. While these datasets are more realistic from both hardware and environment perspectives, it is hard to quantify failure cases accurately (e.g. model fails on a particular walking variation). Smaller, more controlled datasets such as CASIA-B and FVG have a smaller number of identities, but with a fixed number of covariates. PsyMo brings together the best of both worlds by replicating real-world video hardware for capturing walks and annotating each run with more variations and with metadata information about each subject (gender, age, weight, height). In Table \ref{tab:dataset-comparison}, we showcase a comparison of PsyMo to other popular datasets for the gait recognition task. As such, PsyMo can be used as an additional benchmark and drop-in replacement of older datasets (e.g., CASIA-B \cite{yu2006framework}, de facto standard) for evaluating gait recognition models. 

{\renewcommand{\arraystretch}{1.0}
    \begin{table*}[h]
        \small
        \centering
        \caption{Comparison of popular datasets for gait recognition. PsyMo is similar in size and variations to other datasets, but is annotated with gender, height, weight and psychological attributes for each subject. GREW \cite{zhu2021gait}, Gait3D \cite{zheng2022gait} and DenseGait \cite{cosma22gaitformer} are collected in the wild and have no clear delimitation between variations and viewpoints.}
        \resizebox{0.9\linewidth}{!}{
        \begin{tabular}{l|cccp{3.2cm}ccccp{4.0cm}}
        \textbf{Dataset} & \textbf{Type} & \textbf{\# IDs} & \textbf{\# Seq.} & \textbf{Variations} & \textbf{Views} & \textbf{Env.} & \textbf{Demogr.} & \textbf{BMI} & \textbf{Particularity} \\
             \midrule
        % TUM-GAID \cite{hofmann2014tum} & 305 & 3,370 & CB, CL & 1 & Indoor & \xmark & \xmark & no longer available\\
        FVG \cite{fvg} & Controlled & 226 & 2,857 & NM, CL, BG, WS, CBG & 3 & Outdoor & \xmark & \xmark & time difference \\
        CASIA-B \cite{yu2006framework} & Controlled & 124 & 13,640 & NM, CL, BG & 11 & Indoor & \xmark & \xmark & -- \\
        OU-ISIR \cite{Makihara_CVATN2012} & Controlled & 10,307 & 144,298 & - & 14 & Indoor & \cmark & \xmark & treadmill\\
        Gait3D \cite{zheng2022gait}& In the Wild & 4,000 & 25,309 & - & - & Indoor & \xmark & \xmark & collected in a supermarket \\
        GREW \cite{zhu2021gait} & In the Wild & 26,000 & 128,000 & - & - & Outdoor & \cmark & \xmark & -- \\
        DenseGait \cite{cosma22gaitformer} & In the Wild & 217,954 & 217,954 & - & - & Outdoor & \xmark & \xmark & auto labelled with 42 attributes\\
        \midrule
        PsyMo (\textbf{ours}) & Controlled & 312 & 14,976 & NM, CL, BG, WSS, WSF, TXT, PH & 6 & Indoor & \cmark & \cmark & 17 psychological traits
        \end{tabular}
        }
        \label{tab:dataset-comparison}
    \end{table*}
}

Since PsyMo is by design similar in collection and variations with CASIA-B \cite{yu2006framework}, we propose the same evaluation procedures: performing for each variation $6\times6=36$ experiments, for each viewpoint. In each experiment, the gallery set is comprised of only the second NM run, while the probe set is comprised of all other variations and viewpoints, excluding the viewpoint present in the gallery. This protocol evaluates the model's robustness to changes in viewpoints and walking variations.

\section{Baseline Results}
{\renewcommand{\arraystretch}{1.0}
    \begin{table*}[h]
   \small
   \centering
   \caption{Baseline results for each of the three modalities. Results presented are weighted F1 scores, averaged across variations / viewpoints. We present both run-level and subject-level results (i.e. identity-based majority voting).}
    \resizebox{1.0\linewidth}{!}{
   \begin{tabular}{c|ll|ccccc|c|cccc|ccc|ccc|c}
   & \textbf{Method} & \textbf{Modality}
   & \multicolumn{5}{c}{\textbf{BFI} ({\small 4-classes})} 
   & \multicolumn{1}{c}{\textbf{RSE} ({\small 3-classes})}
   & \multicolumn{4}{c}{\textbf{BPAQ} ({\small 4-classes})}
   & \multicolumn{3}{c}{\textbf{OFER} ({\small 4-classes})}
   & \multicolumn{3}{c}{\textbf{DASS}  ({\small 5-classes})}
   & \multicolumn{1}{c}{\textbf{GHQ} ({\small 3-classes})} \\
   & &  & \textbf{O} & \textbf{C} & \textbf{E} & \textbf{A} & \textbf{N} 
   & \textbf{Esteem}
   & \textbf{Phys.} & \textbf{Verbal} & \textbf{Anger} & \textbf{Host.}
   & \textbf{Chronic} & \textbf{Acute} & \textbf{Recov.}
   & \textbf{Depr.} & \textbf{Anxiety} & \textbf{Stress} 
   & \textbf{GHQ} \\
   \midrule
   \parbox[t]{2mm}{\multirow{5}{*}{\rotatebox[origin=c]{90}{\textbf{Run-Level}}}}
    & GaitFormer \cite{cosma22gaitformer} & 2D Skeletons & 23.80 & 25.35 & \textbf{27.88} &  \textbf{28.12} &     28.21 & 43.59 &   28.27 & \textbf{31.89} &   \textbf{31.24} &    28.84 &    27.43 &  33.05 &   34.83 &     28.27 &  24.55 & 43.69 & 49.38 \\
    & GaitGraph \cite{teepe2021gaitgraph} &  2D Skeletons &  22.10 & 24.89 & 26.39 &  26.09 &     23.67 & 45.49 &   27.86 & 24.21 &   24.24 &    27.83 &    25.17 &  29.61 &   30.00 &     30.82 &  19.45 & 42.73 & 52.94 \\
    & GaitSet \cite{chao2019gaitset} &   Silhouettes &  26.60 & \textbf{26.51} & 25.81 &  26.87 &     27.89 & 45.40 &   26.51 & 29.93 &   27.71 &    28.81 &    \textbf{31.13} &  33.14 &   34.80 &     42.20 &  \textbf{27.57} & \textbf{51.98} & \textbf{56.11} \\
    & GaitGL \cite{lin2021gait} &   Silhouettes &  \textbf{26.64} & 23.48 & 25.70 &  25.94 &     28.30 & 44.73 &   26.28 & 29.99 &   28.16 &    27.90 &    30.35 &  31.92 &   32.95 &     37.64 &  25.03 & 46.03 & 54.79 \\
    & SMPLGait \cite{zheng2022gait} & Silhouettes + SMPL &  24.14 & 20.83 & 23.21 &  25.75 &     \textbf{32.21} & \textbf{45.90} &   \textbf{28.32} & 28.09 &   24.13 &    \textbf{30.18} &    30.07 &  \textbf{34.54} &   \textbf{35.05} &     \textbf{42.74} &  21.91 & 51.67 & 52.22 \\    
    
    \midrule
    \parbox[t]{2mm}{\multirow{5}{*}{\rotatebox[origin=c]{90}{\textbf{Subject-Level}}}} 
    & GaitFormer \cite{cosma22gaitformer} &  2D Skeletons &  \textbf{33.12} & 31.03 & \textbf{36.27} &  \textbf{38.42} &     33.25 & \textbf{55.58} &   \textbf{38.00} & \textbf{42.81} &   \textbf{42.73} &    \textbf{39.09} &    \textbf{36.71} &  \textbf{38.88} &   \textbf{46.80} &     38.45 &  \textbf{36.33} & 52.85 & 57.21 \\
    & GaitGraph \cite{teepe2021gaitgraph} &  2D Skeletons &  24.40 & 28.91 & 27.55 &  31.33 &     27.55 & 50.95 &   34.81 & 23.76 &   26.51 &    32.37 &    28.19 &  28.30 &   25.08 &     35.88 &  19.94 & 48.01 & 53.10 \\
   & GaitSet \cite{chao2019gaitset} &   Silhouettes &  27.50 & \textbf{31.55} & 26.55 &  27.13 &     27.37 & 48.38 &   32.37 & 36.46 &   30.94 &    30.87 &    32.99 &  33.13 &   36.70 &     \textbf{46.64} &  33.53 & 55.05 & \textbf{60.59} \\
    & GaitGL \cite{lin2021gait} &   Silhouettes &  31.04 & 23.78 & 28.05 &  23.15 &     30.73 & 48.35 &   26.89 & 41.79 &   30.37 &    28.80 &    34.22 &  36.96 &   37.31 &     46.20 &  27.75 & \textbf{55.57} & 60.54 \\
    & SMPLGait \cite{zheng2022gait} & Silhouettes + SMPL &  22.77 & 18.02 & 19.74 &  26.80 &     \textbf{36.51} & 48.62 &   33.80 & 33.42 &   20.47 &    34.89 &    31.83 &  37.27 &   38.71 &     42.66 &  21.89 & 52.72 & 58.97 \\

   \end{tabular}
    }
   \label{tab:baselines}
    \end{table*}
}
% For psychological trait estimation, we provide baseline results for the three data modalities present in this dataset. Due to space constraints, we showcase baseline results only for trait estimation. In Table \ref{tab:baselines}, we present the weighted F$_1$ score averaged across all walking variations and viewpoints. For 2D skeletons, we used GaitFormer \cite{cosma22gaitformer}, a skeleton transformer model used in gait recognition, and GaitGraph \cite{teepe2021gaitgraph}, a graph-based skeleton model. Further, for silhouettes, we used GaitSet \cite{chao2019gaitset} and GaitGL \cite{lin2021gait}, two models achieving state-of-the-art results for gait recognition. Finally, for SMPL body meshes, we trained SMPLGait \cite{zheng2022gait}, a model combining both silhouettes and SMPL parameters to spatially transform the silouettes according to the camera angle. For each factor / subscale, we train a separate model using a standard weighted cross-entropy loss. Each model is trained three times, and the results are averaged. For \textit{subject-level} evaluation, we make predictions for every run of an identity and the final prediction is the majority vote across runs. 

We provide baseline results for psychological trait estimation for the three data modalities present in this dataset. In Table \ref{tab:baselines}, we present the weighted F$_1$ score averaged across all walking variations and viewpoints. For 2D skeletons, we used GaitFormer \cite{cosma22gaitformer}, a skeleton transformer model used in gait recognition, and GaitGraph \cite{teepe2021gaitgraph}, a graph-based skeleton model. Further, for silhouettes, we used GaitSet \cite{chao2019gaitset} and GaitGL \cite{lin2021gait}, two models achieving state-of-the-art results for gait recognition. Finally, for SMPL body meshes, we trained SMPLGait \cite{zheng2022gait}, a model combining both silhouettes and SMPL parameters to spatially transform the silhouettes according to the camera angle. We train a separate model for each factor / subscale using a standard weighted cross-entropy loss. Each model is trained three times, and the results are averaged. For \textit{subject-level} evaluation, we make predictions for every run of an identity and the final prediction is the majority vote across runs.

Preliminary baseline results show that some psychological factors, such as mental distress and fatigue are more easily discernible from silhouettes than from skeletons. SMPLGait \cite{zheng2022gait}, which incorporates SMPL parameters and silhouette sequences into the model, fairs reasonably well in run-level evaluation compared to other silhouette methods, but does not seem to have a substantial improvement in subject-level. This implies that the predictions are not consistent across the different walks of the same person. GaitFormer \cite{cosma22gaitformer}, the only transformer-based architecture that considers a whole skeleton as a token, has substantial improvements in subject-level evaluation, outperforming the majority of other models across modalities. 

Naturally, the \textit{subject-level} evaluation protocol yields significantly higher results, prompting the future use of a more general class of models that can process sets of runs in an end-to-end fashion. PsyMo is a challenging dataset, and encourages the development of novel architectures that are better able to estimate psychological traits from sets of walking sequences. Moreover, future works might exploit the multi-modal nature of PsyMo, as well as the natural correlations between questionnaire subscales and employ a multi-modal / multi-task approach.

\section{Limitations and Societal Broader Impact}
\label{sec:limitations}
While we strive to make PsyMo realistic from a surveillance hardware standpoint (i.e. commercial cameras, 16 fps, HD resolution), the researcher's presence in the laboratory might affect the manner of walking of the subject. Takayanagi et al. \cite{takayanagi2019relationship} found a weak positive correlation between the number of researchers present in the indoor laboratory and the gait speed of subjects. However, this correlation is significantly weaker when only one researcher is present. Our dataset contains mostly students of similar age, all of whom are from Romania, which makes the dataset biased in this age group, culture and upbringing. As such, the intended usage of PsyMo is to pave the way for exploratory methods into psychological traits present in movement, in the form of walking, and to develop narrow models in this domain. It is not meant to be a benchmark generalizable across all age groups, cultures and races. The recommended approach for this dataset is fine-tuning on PsyMo general-purpose models pretrained on diverse datasets such as GREW \cite{zhu2021gait} or DenseGait \cite{cosma22gaitformer}. In terms of the potential negative societal impact, we acknowledge that gait analysis can be used for intrusive surveillance practices, and advances in gait analysis may allow bad actors to profile and target individuals without their consent. However, PsyMo is intended only for research purposes and only enables the possibility for interdisciplinary studies of psychological manifestations into gait.

\section{Conclusions}
We proposed PsyMo, a dataset containing 14,976 walking sequences captured with commodity video surveillance hardware and annotated with 17 self-reported psychological attributes. It contains 312 subjects walking in different 6 viewpoints and 7 variations, including two actions-while-walking / dual-task variations (texting and talking on the phone). The dataset is anonymized, and we release only processed silhouettes extracted by an HTC model \cite{chen2019hybrid}, skeleton sequences extracted with AlphaPose \cite{li2018crowdpose}, and SMPL body meshes extracted with CLIFF \cite{li2022cliff}. Additional data modalities and updated representations with improved models will be added over time, if necessary. The purpose of PsyMo is to encourage interdisciplinary research into automatic estimation of psychological traits in biomechanical movement. The dataset is challenging and it encourages the development of novel, higher-level architectures that are able to take advantage of multiple walking sequences at the same time. The natural correlations between psychological subscales can be exploited in a multi-task fashion. Due to its size and diversity, PsyMo can also be used to benchmark methods for gait recognition as a drop-in replacement or alongside other popular gait datasets. 

{\small
\subsection*{Acknowledgements}
%We thank Diana Todea for her insightful psychology-related suggestions and discussions. The dataset collection would not have been possible without the help of Andy Catruna, Laura Ruse and Dan Tudose.
This work was partly supported by CRC research grant 2021, with funds from UEFISCDI in project CORNET (PN-III 1/2018) and by the Google IoT/Wearables Student Grants. We thank Diana Todea for her psychology insights, and Andy Catruna, Laura Ruse and Dan Tudose for their help with the dataset collection. 
}

{\small
\bibliographystyle{ieee_fullname}
\bibliography{refs}
}

\appendix
\section*{Datasheet for PsyMo: A Dataset for Estimating Self-Reported Psychological Traits from Gait}
%%%%%%%%%%%%%%%%%%%%%%%%%%%%%%%%%%%%%%%%%%%%%%%%%%%%%%%%%%%%%%%%%%%%%%%%%%%%%%%%
\section{Motivation For Datasheet Creation}

\textcolor{blue}{\subsection{Why was the datasheet created? (e.g., was there a specific task in mind? was there a specific gap that needed to be filled?)}}

This datasheet was created to accompany the WACV 2024 submission for the PsyMo dataset, and details the composition, collection process, distribution, preprocessing, maintanance and ethical considerations for the PsyMo dataset. PsyMo is intended for exploration of psychological manifestations in walking patterns. PsyMo can be used for benchmarking models in the estimation of 17 psychometric attributes from gait in multiple variations.

\textcolor{blue}{\subsection{Has the dataset been used already? If so, where are the results so others can compare
(e.g., links to published papers)?}}

The dataset has not been currently used in other works. We invite researchers to propose baseline results for the psychological trait estimation and gait recognition tasks mentioned in the article.

\textcolor{blue}{\subsection{What (other) tasks could the dataset be used for?}}

PsyMo's main purpose is estimation of psychometric attributes from gait. However, due to its size and controlled diversity of walking variations and viewpoints, it could be used to benchmark models in gait recognition.

\textcolor{blue}{\subsection{Who funded the creation dataset?}}

The dataset creation received no external funding. All subjects in the dataset were volunteers.

\section{Datasheet Composition}
\textcolor{blue}{\subsection{What are the instances?(that is, examples; e.g., documents, images, people, countries) Are there multiple types of instances? (e.g., movies, users, ratings; people, interactions between them; nodes, edges)}}

PsyMo contains 312 different subjects walking captured in different variations and camera viewpoints. We process the walking sequences using a state-of-the-art AlphaPose \cite{li2018crowdpose} to obtain skeleton sequences, CLIFF \cite{li2022cliff} to estimate 3D human pose and 3D meshes in the form of parametric SMPL predictions, and extracted silhouettes using pretrained instance segmentation model (Hybrid Task Cascade \cite{chen2019hybrid}) to obtain silhouette sequences. Skeletons are composed of 18 joint coordinates in the image plane, with x and y coordinates and an additional confidence score for each joint, which measures detection quality. Each sequence is provided in JSON format. SMPL information is provided in .npy files, with the same format as in the opensource implementation \footnote{\url{https://github.com/huawei-noah/noah-research/tree/master/CLIFF}}. Additionally, each silhouette is provided in a 128x128 image in .PNG format. Each silhouette is centered in the frame. We also provide Gait Energy Images (GEI), for convenience, in .PNG format. GEI's are constructed by averaging the silhouettes of a walk.

\textcolor{blue}{\subsection{How many instances are there in total (of each type, if appropriate)?}}
The are a total of 14,976 walking sequences, from 312 individuals. Each subject has 48 walks, across 6 viewpoints (0$^{\circ}$, 45$^{\circ}$, 90$^{\circ}$, 180$^{\circ}$, 225$^{\circ}$, 270$^{\circ}$ including round-trips) and 7 walking variations (normal walking, carrying bag, clothing change, walk speed slow, walk speed fast, talking on the phone and texting while walking). For normal walking, subjects walked a total of 4 times, while on other variations each subject walked 2 times.

\textcolor{blue}{\subsection{What data does each instance consist of ? “Raw” data (e.g., unprocessed text or images)? Features/attributes? Is there a label/target associated with instances? If the instances related to people, are sub-populations identified (e.g., by age, gender, etc.) and what is their distribution?}}

Skeleton sequences and SMPL parameters are not further processed, and are provided as obtained by AlphaPose and CLIFF, respectively. For silhouettes, raw pixel probabilities are provided, but the silhouettes are centered in the image and rescaled to 128x128. Each subject is associated with age, gender, weight and height information, alongside raw scores and ordinal classes for 17 psychological attributes.

\begin{figure*}[hbt!]
    \centering
    \includegraphics[width=\textwidth]{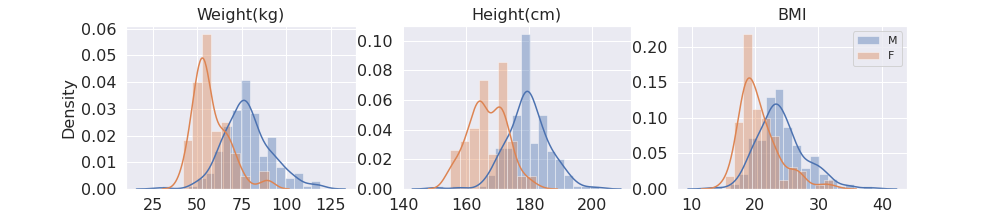}
    \caption{Distributions of personal attributes (weight, height, BMI) for all participants, split by gender.}
    \label{fig:gender-distribution}
\end{figure*}

Our dataset consists of student volunteers. Figure \ref{fig:gender-distribution} showcases the gender distribution across weight, height and BMI.

\textcolor{blue}{\subsection{Is there a label or target associated with each instance? If so, please provide a description.}}

Each walking sequence has information regarding the viewpoint and walking variation. For each subject, there is information regarding their age, gender, weight, height. Additionally, raw scores 6 psychological questionnaires are provided, totalling 17 psychological attributes, across factors and subscales. The questionnaires are the Big Five Index (BFI)\cite{john1999big}, Rosenberg Self-Esteem (RSE) \cite{Rosenberg+2015}, Buss-Perry Aggression Questionnaire (BPAQ) \cite{buss1992aggression}, Ocupational Fatigue Exhaustion / Recovery Scale (OFER) \cite{winwood2005development}, Depression, Anxiety and Stress Scale (DASS) \cite{lovibond1995} and General Health Questionnaire (GHQ) \cite{goldberg1970psychiatric}. We also provide ordinal classes for each subscale / factor. For BFI and BPAQ we obtain ordinal classes using quantiles on the raw score, while on the others we used their respective threshold values.

\textcolor{blue}{\subsection{Is any information missing from individual instances? If so, please provide a description, explaining why this information is missing (e.g., because it was unavailable). This does not include intentionally removed information, but might include, e.g., redacted text.}}

There is no missing information for each instance / subject.

\textcolor{blue}{\subsection{Are relationships between individual instances made explicit (e.g., users’ movie ratings, social network links)? If so, please describe how these relationships are made explicit.}}

There are no relationships between subjects or walking sequences. Each walking sequence has explicit information regarding viewpoint and walking variation.

\textcolor{blue}{\subsection{Does the dataset contain all possible instances or is it a sample (not necessarily random) of instances from a larger set? If the dataset is a sample, then what is the larger set? Is the sample representative of the larger set (e.g., geographic coverage)? If so, please describe how this representativeness was validated/verified. If it is not representative of the larger set, please describe why not (e.g., to cover a more diverse range of instances, because instances were withheld or unavailable).}}

PsyMo does not cover all possible viewpoints and walking variations, and is instead collected in a controlled environment to benchmark performance on specific variations, available for all subjects in the dataset, similar to other datasets in this domain \cite{yu2006framework,fvg,ouisir}. Datasets such as DenseGait \cite{cosma22gaitformer} and GREW \cite{zhu2021gait} provide ample diversity of walking variations present in the real world, but fine-grained annotation with psychological attributes is unfeasible. All subjects contained in PsyMo are Romanian, however it is a representative sample for a proof of concept on estimating psychometric attributes from walking. 

\textcolor{blue}{\subsection{Are there recommended data splits (e.g., training, development/validation, testing)? If so, please provide a description of these splits, explaining the rationale behind them.}}

For all tasks, we recommend a 80:20 training / validation split on the subjects, corresponding to 250 subject for training and 62 subject for validation.

\textcolor{blue}{\subsection{Are there any errors, sources of noise, or redundancies in the dataset? If so, please provide a description.}}

The skeleton provided by the pose estimation model might not always be correctly extracted, due to occlusion of joints in some viewpoints. However, we did not address this issue (for example, by using dedicated hardware for skeleton estimation such as Kinect) as this would not be available in realistic surveillance scenarios. As such, approaches that process PsyMo for a particular task may also include a way to rectify skeletons. The same is true for the extracted silhouettes and SMPLs.

\textcolor{blue}{\subsection{Is the dataset self-contained, or does it link to or otherwise rely on external resources (e.g., websites, tweets, other datasets)? If it links to or relies on external resources, a) are there guarantees that they will exist, and remain constant, over time; b) are there official archival versions of the complete dataset (i.e., including the external resources as they existed at the time the dataset was created); c) are there any restrictions (e.g., licenses, fees) associated with any of the external resources that might apply to a future user? Please provide descriptions of all external resources and any restrictions associated with them, as well as links or other access points, as appropriate.}}

The dataset is self-contained.

\section{Collection Process}

\textcolor{blue}{\subsection{What mechanisms or procedures were used to collect the data (e.g., hardware apparatus or sensor, manual human curation, software program, software API)? How were these mechanisms or procedures validated?}}

We used three Tapo C200 \footnote{\url{https://www.tp-link.com/ro/home-networking/cloud-camera/tapo-c200/}} consumer surveillance cameras, for their ease of use, widespread availability and web API \footnote{\url{https://github.com/JurajNyiri/pytapo}}. The cameras was synchronized and controlled using a custom built python program. Subjects filled in 6 psychological questionnaires remotely using Google Forms, alongside attributes related to their body composition (age, gender, weight, height). Questionnaires are validated on large-scale populations, and have official translations in Romanian; if there are no translations, we translated them with the help of a psychologist.

\textcolor{blue}{\subsection{How was the data associated with each instance acquired? Was the data directly observable (e.g., raw text, movie ratings), reported by subjects (e.g., survey responses), or indirectly inferred/derived from other data (e.g., part-of-speech tags, model-based guesses for age or language)? If data was reported by subjects or indirectly inferred/derived from other data, was the data validated/verified? If so, please describe how.}}

\begin{figure}[hbt!]
    \centering
    \includegraphics[width=0.95\linewidth]{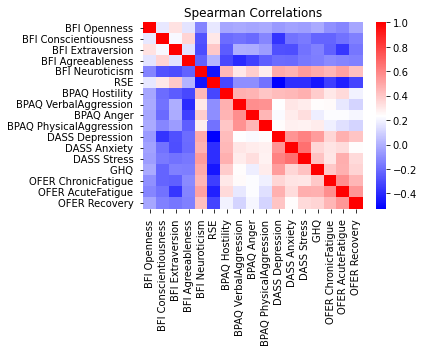}
    \caption{Spearman correlations between questionnaire scores.}
    \label{fig:ghq-dass}
\end{figure}

The data associated with each instance is self-reported individually by each subject through survey responses. Questionnaires have been validated in literature and contain redundancies both in terms of individual items in each questionnaire. Moreover, we have some redundancies / correlations across questionnaires, see Figure \ref{fig:ghq-dass})

\textcolor{blue}{\subsection{If the dataset is a sample from a larger set, what was the sampling strategy (e.g., deterministic, probabilistic with specific sampling probabilities)?}}

The dataset is not sampled from a larger set.

\textcolor{blue}{\subsection{Who was involved in the data collection process (e.g., students, crowdworkers, contractors) and how were they compensated (e.g., how much were crowdworkers paid)?}}

The dataset is comprised of student volunteers.

\textcolor{blue}{\subsection{Over what timeframe was the data collected? Does this timeframe match the creation timeframe of the data associated with the instances (e.g., recent crawl of old news articles)? If not, please describe the timeframe in which the data associated with the instances was created.}}

The dataset was collected over the course of 2 months. Each subject was required to fill in the psychological questionnaires and then to present themselves to the lab to have their walk captured. 
% This timeframe validates all the questionnaire requirements, as for example, the DASS scale has questions referring to states of mind from the past week.

\section{Data Preprocessing}

\textcolor{blue}{\subsection{Was any preprocessing/cleaning/labeling of the data done (e.g., discretization or bucketing, tokenization, part-of-speech tagging, SIFT feature extraction, removal of instances, processing of missing values)? If so, please provide a description. If not, you may skip the remainder of the questions in this section.}}

There has not been any significant preprocessing done on the dataset, except for the calculation of the final scores on each questionnaires, according to the respective specification for each questionnaire.

\textcolor{blue}{\subsection{Was the "raw" data saved in addition to the preprocessed/cleaned/labeled data (e.g., to support unanticipated future uses)? If so, please provide a link or other access point to the "raw" data.}}

N/A

\textcolor{blue}{\subsection{Is the software used to preprocess/clean/label the instances available? If so, please provide a link or other access point.}}

N/A

\textcolor{blue}{\subsection{Does this dataset collection/processing procedure achieve the motivation for creating the dataset stated in the first section of this datasheet? If not, what are the limitations?}}

N/A

\textcolor{blue}{\subsection{Any other comments}}

\section{Dataset Distribution}

\textcolor{blue}{\subsection{How will the dataset be distributed? (e.g., tarball on website, API, GitHub; does the data have a DOI and is it archived redundantly?)}}

Currently, the dataset is distributed at \url{https://bit.ly/3Q91ypD}.

\textcolor{blue}{\subsection{When will the dataset be released/first distributed? What license (if any) is it distributed under?}}

The dataset is available immediately at \url{https://bit.ly/3Q91ypD}. The dataset is protected by the CC-BY-NC-ND\footnote{\url{https://creativecommons.org/licenses/by-nc-nd/4.0/legalcode}} License.

\textcolor{blue}{\subsection{Are there any copyrights on the data?}}

The dataset does not contain any copyrighted content, and was collected entirely by the authors. The dataset is protected by copyright through the CC-BY-NC-ND License.

\textcolor{blue}{\subsection{Are there any fees or access/export restrictions?}}

There are no fees or restrictions.

\section{Dataset Maintenance}

\textcolor{blue}{\subsection{Who is supporting/hosting/maintaining the dataset?}}

The dataset is supported, hosted and maintained by the authors. 
% It will be hosted on a public URL on a subdomain of \verb|upb.ro|

\textcolor{blue}{\subsection{Will the dataset be updated? If so, how often and by whom?}}

If there is a rationale for updating the dataset (e.g. extending or correcting it), the authors will make the necessary modifications.

\textcolor{blue}{\subsection{How will updates be communicated? (e.g., mailing list, GitHub)}}

Updates will be posed on the hosting website.

\textcolor{blue}{\subsection{If the dataset becomes obsolete how will this be communicated?}}

In case the dataset will become obsolete, it will be rectracted and an update will be posted on the hosting website.

\textcolor{blue}{\subsection{Is there a repository to link any/all papers/systems that use this dataset?}}

We will make a dedicated webpage on the hosting website which will feature any system that uses PsyMo.

\textcolor{blue}{\subsection{If others want to extend/augment/build on this dataset, is there a mechanism for them to do so? If so, is there a process for tracking/assessing the quality of those contributions. What is the process for communicating/distributing these contributions to users?}}

N/A. Currently PsyMo is not intended to be extended by third parties except the authors.

\section{Legal and Ethical Considerations}

\textcolor{blue}{\subsection{Were any ethical review processes conducted (e.g., by an institutional review board)? If so, please provide a description of these review processes, including the outcomes, as well as a link or other access point to any supporting documentation.}}

% The collection of PsyMo has been approved by the Ethics Review Board of the Faculty of Computer Science at University Politehnica of Bucharest. The approval documentation can be found at \url{bit.ly/3MXSgKz}. The documentations for the approval includes a complete description of the data collection procedure and can be found at \url{bit.ly/3Qnc4cZ}.

The collection of PsyMo has been approved by the Ethics Review Board. The approval documentation and a complete description of the data collection procedure will be made available after the anonymization period.

% can be found at \url{bit.ly/3MXSgKz}. The documentations for the approval includes  and can be found at \url{bit.ly/3Qnc4cZ}.

\textcolor{blue}{\subsection{Does the dataset contain data that might be considered confidential (e.g., data that is protected by legal privilege or by doctor patient confidentiality, data that includes the content of individuals non-public communications)? If so, please provide a description.}}

No, the dataset does not contain confidential information. All data is self-reported by each participant under explicit and informed consent.

\textcolor{blue}{\subsection{Does the dataset contain data that, if viewed directly, might be offensive, insulting, threatening, or might otherwise cause anxiety? If so, please describe why}}

No, PsyMo does not contain offensive information, it is comprised of walking sequences (skeletons and silhouettes), annotated with 17 psychometric attributes.

\textcolor{blue}{\subsection{Does the dataset relate to people? If not, you may skip the remaining questions in this section.}}

Yes, the dataset is comprised of walking sequences from 312 volunteer subjects.

\textcolor{blue}{\subsection{Does the dataset identify any subpopulations (e.g., by age, gender)? If so, please describe how these sub-populations are identified and provide a description of their respective distributions within the dataset.}}

Of the 312 participants, 113 were female and 199 were male, with an average age of 21.9 years (SD = 2.18). Moreover, the average weight for the participants is 70.5kg (SD = 15.7) with the average height of 174.8cm (SD = 8.9), corresponding to an average BMI of 22.87 (SD = 3.9). 

\textcolor{blue}{\subsection{Is it possible to identify individuals (i.e., one or more natural persons), either directly or indirectly (i.e., in combination with other data) from the dataset? If so, please describe how.}}

The dataset is anonymized: we do not release the raw walking videos, only anonymized skeleton sequences and silhouette sequences. It is not possible to identify subjects unless their walking sequence is annotated with their identity.

\textcolor{blue}{\subsection{Does the dataset contain data that might be considered sensitive in any way (e.g., data that reveals racial or ethnic origins, sexual orientations, religious beliefs, political opinions or union memberships, or locations; financial or health data; biometric or genetic data; forms of government identification, such as social security numbers; criminal history)? If so, please provide a description.}}

The dataset does not contain any data related to the ones enumerated above. PsyMo contains responses from 6 psychological questionnaires, two of which related to mental health (DASS-21 and GHQ-9). However, they were provided under explicit and informed consent, and any identifiable information has been removed; PsyMo is anonymized.

\textcolor{blue}{\subsection{Did you collect the data from the individuals in question directly, or obtain it via third parties or other sources (e.g., websites)?}}

All data was directly self-reported by each participant. Each walking instance is directly captured with explicit and informed consent in predetermined variations.

\textcolor{blue}{\subsection{Were the individuals in question notified about the data collection? If so, please describe (or show with screenshots or other information) how notice was provided, and provide a link or other access point to, or otherwise reproduce, the exact language of the notification itself.}}

We announced our intention of collecting this dataset with the following message:

\textit{"This assignment represents an opportunity for you to contribute to the research performed in our Computer Science department of our university, by helping us collect a dataset which will enable an interdisciplinary study on personality traits and movement. This requires you filling in 6 personality questionnaires and walking multiple times in front of three cameras."}

Students voluntarily participated in this study in their own terms, with full knowledge of the dataset collection procedure, distribution and intended purposes. The dataset collection was approved by the Ethical Review Board.

% at University Politehnica of Bucharest (see \url{bit.ly/3MXSgKz})

\textcolor{blue}{\subsection{Did the individuals in question consent to the collection and use of their data? If so, please describe (or show with screenshots or other information) how consent was requested and provided, and provide a link or other access point to, or otherwise reproduce, the exact language to which the individuals consented.}}

Before filling in the questionanires and having their walk captured, subjects were prompted for their consent, after a description of the study (translated to English from Romanian):

\textit{This research is carried out within the Department of Computers Science. Your participation will help us explore the possible correlations between psychometric information and human movement in physical space. This is a pilot study of an unexplored area in the field of psychology and artificial intelligence. The aim of this study is to investigate the manifestations of psychometric attributes on gait in different scenarios that simulate real gait situations (normal gait, different clothing, backpack etc).The study consists of completing a set of psychometric questionnaires that measure a set of attributes related to personality and psychological disorders. After completing the questionnaires, your walking patterns will be recorded in different scenarios after an appointment.}

\textit{In accordance with the requirements of Regulation (EU) 2016/679 on the protection of individuals with regard to the processing of personal data and on the free movement of such data and repealing Directive 95/46/EC (General Data Protection Regulation) and Law no. 506/2004 on the processing of personal data and the protection of privacy, the research team has the obligation to manage safely and only for the specified purposes. The data you will provide: demographic data, answers to questionnaires and movement information.
The statistical processing of the provided data will be analyzed at the sample level and will not be presented individually in any scientific publication. The recorded information will only be used by members of the research team.
After the collection process is completed, the data will be anonymized, of interest being only the movement information. The results of the research will be made public only in an anonymized version, without being able to reach the identity of the people present in the study.
Your participation in this research is entirely voluntary. By choosing to participate you agree to the processing of personal data provided for research purposes.
If you have any questions, concerns, or complaints, or if you would like to report any research-related harm or abuse, please contact us.}

\textcolor{blue}{\subsection{If consent was obtained, were the consenting individuals provided with a mechanism to revoke their consent in the future or for certain uses? If so, please provide a description, as well as a link or other access point to the mechanism (if appropriate).}}

Subjects can revoke their consent by directly contacting the authors via email.

\textcolor{blue}{\subsection{Has an analysis of the potential impact of the dataset and its use on data subjects (e.g., a data protection impact analysis) been conducted? If so, please provide a description of this analysis, including the outcomes, as well as a link or other access point to any supporting documentation.}}

A data protection impact analysis has not been performed on PsyMo. The dataset does not contain sensitive information and is anonymized.

\end{document}